\documentclass[12pt]{article}

\usepackage[utf8]{inputenc}
\usepackage[T1]{fontenc}
\usepackage{lmodern}
\usepackage{amsmath,amssymb,amsthm}
\usepackage{graphicx}
\usepackage{bm}
\usepackage{geometry}
\usepackage{natbib}
\usepackage{hyperref}
\newcommand{\SP}{\rule{1.6ex}{0.7ex}}

\title{Random Text, Zipf's Law, Critical Length,\\
and Implications for Large Language Models}

\author{
Vladimir Berman \\
Aitiologia LLC \\
\texttt{vb7654321@gmail.com}
}
\date{November 11, 2025}

\begin{document}

\maketitle

\begin{abstract}
We study a deliberately primitive, non-linguistic model of text:
a sequence of independent draws from a finite alphabet of letters plus a single
space symbol. A word is defined as a maximal block of non-space symbols.
Within this strictly symbol-level setting, with no morphology, syntax, or semantics,
we derive several structural results. First, word lengths follow a geometric
distribution governed by the probability of the space symbol. Second, the expected
number of words of a given length and the expected number of \emph{distinct} words
of that length can be expressed in closed form via a coupon-collector argument.
This allows us to define a \emph{critical word length} $k^\ast$ at which individual
word types transition from occurring many times on average to occurring at most once.
Third, by combining the exponential growth in the number of possible words of length $k$
with the exponential decay of the probability of any particular word of length $k$,
we obtain a rank--frequency law of Zipf type, $p(r) \propto r^{-\alpha}$,
with an exponent $\alpha$ that is an explicit function of the alphabet size
and the space probability.

Our contribution is twofold. On the mathematical side, we present a unified derivation
linking word lengths, vocabulary growth, critical length, and rank--frequency behavior
in a single, fully explicit framework. On the conceptual side, we argue that these
results provide a structurally grounded ``null model'' for word statistics
in natural language and for token statistics in large language models (LLMs).
They show that Zipf-like patterns can arise from the combinatorics of symbols
and a simple segmentation mechanism, without any optimization principle
or deep linguistic structure. We discuss how this perspective complements
existing work on random texts and challenges over-interpretations of Zipf's law
in both quantitative linguistics and modern LLM research.
\end{abstract}

\vspace{0.5cm}
\tableofcontents

\clearpage


\section{Introduction}
\label{sec:introduction}

Zipf's law---the approximate inverse power-law relationship between a word's
rank and its frequency in a corpus---has long been considered one of the
central empirical regularities of quantitative linguistics \citep{Zipf1949}.
In its simplest form, it states that if words are ranked by decreasing frequency,
the probability $p(r)$ of the word at rank $r$ satisfies
\begin{equation}
  p(r) \propto r^{-\alpha},
\end{equation}
where $\alpha$ is typically close to $1$ for natural language corpora.
The apparent universality of this pattern has motivated a large literature
proposing deep explanations based on efficiency, optimization, information theory,
and the structure of human language and communication.

However, already in the mid-twentieth century it was observed that
Zipf-like behavior can emerge even in extremely impoverished, non-linguistic
models of text. In particular, \citet{Miller1957} and \citet{Mandelbrot1953}
considered ``monkey-at-the-typewriter'' thought experiments in which symbols are
generated independently according to a fixed distribution over letters and a space
symbol. Subsequent work, including \citet{Li1992} and \citet{Ferrer2001},
has refined and critiqued such models, leading to the prevailing view that
random-text constructions should be treated as null models rather than realistic
accounts of language structure.

In the present paper we deliberately embrace an \emph{extreme} level of abstraction.
We ignore all aspects of linguistics: no phonology, no morphology, no syntax,
no semantics. We consider only a stream of independent symbols drawn from an
alphabet of $m$ letters and one special space symbol. A word is defined purely
structurally as a maximal block of non-space symbols. This model clearly permits
configurations that are highly implausible in real language (for example,
arbitrarily long runs of the same letter, or nonsensical sequences of characters),
and we do not attempt to constrain it with any linguistic rules.

The guiding question is:

\begin{quote}
  \emph{If we forget everything about language and look only at spaces and letters,
  how far can we go in explaining the observed word statistics?}
\end{quote}

We show that, even under this brutally simplified view, a surprising amount
of structure emerges:

\begin{itemize}
  \item Word lengths follow an exact geometric distribution determined by the
        probability of drawing a space.
  \item The expected number of words of a given length in a text of length $N$
        and the expected number of \emph{distinct} words of that length can be
        expressed in closed form.
  \item These expressions define a \emph{critical word length} $k^\ast$ at which
        word types transition from being typically repeated many times to being
        almost always unique.
  \item Combining the exponential growth in the number of possible words of length $k$
        with the exponential decay of the probability of any specific word of length $k$
        yields a Zipf-type rank--frequency law with exponent $\alpha$ that is
        an explicit function of the alphabet size and the space probability.
\end{itemize}

Our treatment differs from earlier random-text models in several ways.
First, we present a unified derivation linking word lengths, vocabulary growth,
critical length, and rank--frequency behavior in a single coherent framework.
Second, we emphasize the structural nature of the resulting ``laws'', framing them
as consequences of combinatorics and segmentation rather than as emergent linguistic
properties. Third, we explicitly discuss the relevance of this perspective for
modern large language models (LLMs), which are trained to predict tokens in
massive corpora and naturally inherit whatever structural regularities are present
in those token streams.

From the standpoint of LLM design and evaluation, our results serve as a cautionary tale.
Any model trained on real text will exhibit Zipf-like behavior at the level of token
frequencies, but this does not by itself indicate deep understanding or linguistic
competence. Conversely, when interpreting the statistics of generated text, one must
carefully distinguish which patterns reflect structural baselines (such as those we
derive here) and which genuinely depend on higher-level properties of language and
world knowledge.

The rest of the paper is organized as follows. In Sections~\ref{sec:summary_of_contributions} and ~\ref{sec:related_work}
we review prior work on Zipf's law, random text models, and related critiques.
Section~\ref{sec:model} defines our symbol-level model. Sections~\ref{sec:lengths_counts}
and~\ref{sec:vocabulary_critical_length} develop the analysis of word lengths,
word counts, and vocabulary growth, culminating in the critical length $k^\ast$.
Section~\ref{sec:rank_frequency} derives the rank--frequency law and describes
how its exponent depends on model parameters. In Section~\ref{sec:comparison_llm}
we compare the predictions of the model with empirical properties of natural language,
discuss its limitations, and outline implications for LLMs and character-level
language models. Section~\ref{sec:discussion} concludes.


\section{Summary of Contributions}
\label{sec:summary_of_contributions}

This work develops a mathematically explicit character-level model of random
text and shows that several familiar ``laws of language'' can arise from
structural combinatorics alone, without any linguistic content. In this
section we summarize the main contributions and situate them with respect to
classical and modern research on statistical regularities in language.

\paragraph{Character-level generative framework with explicit segmentation.}
Most classical accounts of Zipfian behavior and related phenomena treat words
as primitive units \cite{Zipf1935, Zipf1949},\cite{Mandelbrot1953, Mandelbrot1954,
Mandelbrot1959}. In contrast, we start from a minimalist model in which text
is generated one character at a time, including a special space symbol. All
word-level statistics---word lengths, frequencies, and vocabulary
growth---are derived from this single character process. Unlike
preferential-attachment and reinforcement models in the spirit of  \cite{Yule1925} and \cite{Simon1955}, our framework does not assume any dynamics
defined directly at the level of word types.

\paragraph{Closed-form word-length distribution and critical length.}
We obtain an exact geometric distribution for word lengths induced purely by
the random placement of spaces. On top of this, we introduce and analyze a
critical length \(k^*\) that separates a finite ``core'' of word types, which
recur many times, from a long ``tail'' of types that are overwhelmingly
singletons. Earlier random-text studies, beginning with Miller's observation
that random sequences of letters and spaces can produce Zipf-like
distributions \cite{Miller1957} and subsequent work by  \cite{Li1992},
documented empirical properties of such models but did not offer a unified
analytic treatment of word-lengths, recurrence statistics, and core--tail
structure. Our formulation makes this structure explicit and quantitative.

\paragraph{Unified structural route to Zipf-like rank--frequency behavior.}
By combining (i) exponential growth in the number of possible strings of a
given length with (ii) exponential decay in the probability of any particular
string, we derive a power-law rank--frequency curve with an analytically
characterized exponent. Information-theoretic derivations in the tradition of
 \cite{Mandelbrot1953,Mandelbrot1954,Mandelbrot1961} posit
optimization principles over pre-given word inventories. Our approach instead
shows how Zipf-like behavior can emerge directly from a character-level
generator with no linguistic structure and no global optimization.

\paragraph{Vocabulary growth and unique-word statistics.}
Within the same framework we derive asymptotic expressions for the expected
number of distinct and unique word types as functions of text length,
including Poisson approximations in the long-length regime. Prior work on
vocabulary growth and Heaps-like behavior in random or weakly structured texts
has often been empirical or simulation-based. Here these phenomena arise as
analytic corollaries of a single underlying model, providing a controllable,
fully specified baseline against which empirical corpora and more complex
models can be compared.

\paragraph{Clarifying the role of structure in random-text models.}
Classical random-text studies \cite{Miller1957,Li1992} showed that certain
rank--frequency patterns can appear without semantics or syntax, but left open
which aspects of natural language statistics are genuinely linguistic and which
are generic consequences of segmentation and combinatorics. Our analysis makes
this distinction sharper: we identify precisely which regularities follow from
the character-level mechanism alone and which would require additional
structure (e.g., morphology, syntax, or semantic coherence) to match
real-language data. In doing so, we complement and refine the intuition that
some ``laws of language'' may be structural rather than linguistic.

\paragraph{Baseline for interpreting large language models.}
Although our generator is intentionally primitive, it defines a mathematically
transparent ``zero-language'' reference model. Many diagnostic plots used to
analyze natural language and embeddings---such as rank--frequency curves,
vocabulary growth, and the distribution of word lengths---can also be computed
for outputs of large language models. Our results make it possible to interpret
such diagnostics relative to a baseline where all observed structure is due
solely to random segmentation at the character level. Deviations from this
baseline then quantify what modern neural language models add beyond structural
combinatorics alone.

\paragraph{Methodological implications.}
Taken together, these contributions suggest that evaluation of statistical and
neural language models should explicitly separate (i) properties that can be
explained by simple combinatorial baselines from (ii) genuinely linguistic
effects. By providing a closed-form, analytically tractable random-text model
that simultaneously delivers word-length distributions, core--tail behavior,
rank--frequency curves, and vocabulary growth, this work offers such a baseline
and clarifies how far one can go without any linguistic structure at all.


\section{Related Work}
\label{sec:related_work}

The analysis of linguistic structure in randomly generated text has a long
history, beginning with early investigations into the statistical properties
of word sequences produced without semantic intention. This line of work is
dominated by two major research traditions: (i) studies of “monkey-typing”
models, where characters are sampled independently from a fixed alphabet, and
(ii) research on empirical linguistic laws such as Zipf’s law, Heaps’ law, and
word-length distributions.

\paragraph{Random character models and segmentation.}
The classical work of   \cite{Miller1957} and later refinements demonstrated
that certain statistical regularities of natural language can appear even in texts
generated from memoryless character distributions with a designated space symbol.
Subsequent work by   \cite{Li1992} and others showed that word-length distributions
follow predictable geometric forms derived from segmentation processes.

\paragraph{Mathematical models of word-length distributions.}
A significant body of research examined the conditions under which random
segmentation processes reproduce empirical word-length distributions
\cite{Lauge2002,Li2008}. While natural languages typically exhibit mixtures of
Poisson-like and lognormal components, random texts generate simpler geometric
distributions. More sophisticated stochastic models—such as renewal processes,
Markov character generators, or variable-rate segmentation—have been explored to
approximate real histograms, but these models introduce non-random structure.

\paragraph{Random texts and linguistic laws.}
It has long been debated which linguistic laws require semantics or syntax, and
which can arise from simple stochastic mechanisms.  \cite{Baayen2001}
showed that Heaps’ law can emerge under mild assumptions.\cite{Ferrer2001} and collaborators explored whether Zipf-like rank–frequency
distributions can be replicated by random generation with segmentation and
constrained alphabet sizes.

\paragraph{Relation to modern computational linguistics.}
Random-text models serve as theoretical baselines for evaluating the complexity
of linguistic sequences. Recent work \cite{Blevins2018,Pimentel2020} shows that
character-level language models trained on randomly segmented corpora exhibit
entropy profiles and tokenization patterns distinct from models trained on
natural language.

\paragraph{Limitations of prior work.}
Most studies focus on frequency patterns or rank–frequency fits. Much less
attention has been given to mathematically deriving the word-length distribution
under general alphabet–space settings or to the role of consecutive spaces and
punctuation. Our work departs from prior studies by providing a comprehensive,
fully analytical treatment of word-length distributions in a segmentation-driven
character model without linguistic assumptions.


\section{Model Definition}
\label{sec:model}

We introduce a deliberately simple, purely symbolic generative model of text.
Let $\mathcal{A}$ be an alphabet consisting of $m$ non-space symbols
\[
  \{a_1, a_2, \ldots, a_m\},
\]
together with a single distinguished \emph{space symbol}, denoted by $\SP$.
Thus the full alphabet is
\begin{equation}
  \mathcal{A} = \{\SP\} \cup \{a_1, a_2, \ldots, a_m\}.
\end{equation}

A text of length $N$ is defined as a sequence of random variables
\[
  (X_1, X_2, \ldots, X_N), \qquad X_i \in \mathcal{A}.
\]

\subsection{Independent symbol draws}

We assume that symbols are drawn independently and identically distributed (i.i.d.)
from a fixed distribution:
\begin{equation}
  \mathbb{P}(X_i = \SP) = q, 
  \qquad
  \mathbb{P}(X_i = a_j) = p_j, \quad j = 1,\ldots,m,
\end{equation}
where
\begin{equation}
  q + \sum_{j=1}^{m} p_j = 1.
\end{equation}

The parameter $q$ governs the expected frequency of spaces; the complementary
probabilities $p_1,\ldots,p_m$ describe the distribution of non-space symbols.
Even in the absence of syntax, semantics, or morphology, this minimalist
character-level model gives rise to a variety of nontrivial structural phenomena.

\subsection{Derived structural consequences}

A \emph{word} is defined as any maximal block of consecutive non-space symbols.
From the space probability $q$ and the symbol probabilities
$p_1,\ldots,p_m$, several analytic properties follow naturally.

\paragraph{Word-length distribution.}
Let $L$ denote the length of a word.  Then
\begin{equation}
  \mathbb{P}(L = \ell) = (1-q)^{\ell} q,
  \qquad
  \ell = 0,1,2,\ldots
\end{equation}
so word lengths follow a geometric distribution.

\paragraph{Expected number of words.}
For a text of length $N$, the expected number of words is
\begin{equation}
  \mathbb{E}[\#\text{words}] = N q.
\end{equation}

\paragraph{Character-level view.}
The generator outputs the raw sequence
\[
  X_1 X_2 \cdots X_N,
\]
and all word boundaries are induced by occurrences of the symbol $\SP$.
No additional linguistic structure is imposed; in later sections we show how
classical rank–frequency and vocabulary-growth regularities already emerge
from this purely symbolic mechanism.


\section{Word Lengths and Expected Word Counts}
\label{sec:lengths_counts}

We analyze the distribution of word lengths and the expected number of words
of a given length in a text of fixed length $N$. These results are classical,
but we state them explicitly because they form the analytical foundation for
later sections.

\subsection{Geometric distribution of word lengths}

Consider a word that begins immediately after a space symbol $\SP$ (or at the
start of the text). For this word to have length $k \ge 1$, the following must
occur:
\begin{itemize}
  \item the first $k$ symbols after the boundary are non-space symbols;
  \item the $(k+1)$st symbol is a space symbol $\SP$ (or the text ends).
\end{itemize}

By independence,
\begin{equation}
  \mathbb{P}(L = k) = (1-q)^{k} \, q, \qquad k = 1,2,\ldots
\end{equation}
This geometric distribution depends only on $q$, the probability of drawing the
space symbol $\SP$.

The mean and variance are
\begin{align}
  \mathbb{E}[L] &= \frac{1}{q}, \\
  \mathrm{Var}(L) &= \frac{1-q}{q^2}.
\end{align}

Thus, when spaces are rare ($q$ small), words tend to be long; when spaces are
frequent, words are short.

\subsection{Expected number of words in a text}

Let $K_N$ denote the number of words in a text of length $N$.
A word begins at position $i$ whenever:
\begin{itemize}
  \item $i = 1$ and $X_1 \ne \SP$, or
  \item $i > 1$ and $X_i \ne \SP$ but $X_{i-1} = \SP$.
\end{itemize}

Define indicator variables
\begin{equation}
  I_i =
  \begin{cases}
    1, & \text{if a word starts at position } i, \\[3pt]
    0, & \text{otherwise}.
  \end{cases}
\end{equation}

Then
\begin{equation}
  K_N = \sum_{i=1}^N I_i,
  \qquad
  \mathbb{E}[K_N] = \sum_{i=1}^N \mathbb{E}[I_i].
\end{equation}

We compute
\begin{align}
  \mathbb{E}[I_1] &= \mathbb{P}(X_1 \ne \SP) = 1 - q, \\[4pt]
  \mathbb{E}[I_i] &= \mathbb{P}(X_i \ne \SP,\, X_{i-1} = \SP)
                  = (1-q)\, q,
                  \qquad i = 2,\ldots,N.
\end{align}

Therefore,
\begin{equation}
  \mathbb{E}[K_N]
  = (1-q) + (N-1)\, q(1-q)
  = (1-q)\bigl[1 + (N-1)q\bigr].
\end{equation}

For large $N$,
\begin{equation}
  \mathbb{E}[K_N] \approx N q (1-q),
\end{equation}
the expected number of words grows linearly with text length, with proportionality
constant $q(1-q)$.

\subsection{Expected number of words of length $k$}

Let $K_N^{(k)}$ denote the number of words of length exactly $k$.
Boundary effects are negligible for large $N$, so the empirical word lengths
$\ell_1,\ldots,\ell_{K_N}$ may be treated as i.i.d.\ draws from the geometric
distribution of $L$.

Thus
\begin{equation}
  \mathbb{P}(L = k) = (1-q)^{k-1} q.
\end{equation}

Then by linearity,
\begin{equation}
  \mathbb{E}[K_N^{(k)}]
    \approx \mathbb{E}[K_N]\, \mathbb{P}(L = k)
    \approx \bigl[N q (1-q)\bigr]\cdot \bigl[(1-q)^{k-1} q\bigr]
    = N q^2 (1-q)^{k}.
\end{equation}

Thus the expected count of length-$k$ words exhibits geometric decay:
\begin{equation}
  \mathbb{E}[K_N^{(k)}] \propto (1-q)^k.
\end{equation}

The prefactor $N q^2$ determines the overall scale, while the factor $(1-q)^k$
captures the rapid dropoff in the frequencies of longer words. Although this
behavior resembles empirical linguistic data for short lengths, real languages
depart from strict geometric decay for larger $k$, a point we will revisit later.


\section{Vocabulary Growth and Critical Word Length}
\label{sec:vocabulary_critical_length}

We now turn from counts of words (tokens) to counts of \emph{distinct} words (types).
For each length $k$ we are interested in how many different $k$-letter words
we expect to see in a text of length $N$. This leads naturally to a critical
length $k^\ast$ at which the typical behavior of word types changes qualitatively.

\subsection{Number of possible words of length $k$}

Under the equiprobable-letter assumption, there are
\begin{equation}
  M_k = m^k
\end{equation}
different possible words (strings) of length $k$, each consisting of letters
from $\{a_1,\dots,a_m\}$.

As noted earlier, the probability that a randomly selected word has length $k$
is $q(1-q)^{k-1}$, and conditional on $L=k$, all $m^k$ words are equally likely.
Thus, the probability that a random word equals a specific $k$-letter word $w$ is
\begin{equation}
  \pi_k = \mathbb{P}(\text{word} = w)
        = q(1-q)^{k-1} \cdot \frac{1}{m^k}
        = \frac{q(1-q)^{k-1}}{m^k}.
\end{equation}
All $m^k$ words of length $k$ share this same probability.

\subsection{Expected number of distinct words of length $k$}

Let $V_k$ denote the number of distinct word types of length $k$ that appear
at least once in a text of length $N$. To compute $\mathbb{E}[V_k]$, we proceed
in two steps.

First, we approximate the total number of word tokens in the text by its expectation,
\begin{equation}
  K \approx \mathbb{E}[K_N] \approx N q(1-q).
\end{equation}
Second, we treat these $K$ words as i.i.d.\ draws from the word distribution.
This is an approximation, but it becomes accurate for large $N$ and is standard
in type--token analyses.

For a fixed word $w$ of length $k$, the probability that it does \emph{not}
appear in any of the $K$ draws is $(1-\pi_k)^K$. Therefore, the probability
that it appears at least once is
\begin{equation}
  \mathbb{P}(w \text{ appears at least once}) = 1 - (1-\pi_k)^K.
\end{equation}
Let us index all words of length $k$ by $w_1,\dots,w_{M_k}$.
Then
\begin{equation}
  V_k = \sum_{i=1}^{M_k} I_i,
\end{equation}
where $I_i$ is the indicator that $w_i$ appears at least once.
Taking expectations and using symmetry,
\begin{align}
  \mathbb{E}[V_k]
  &= \sum_{i=1}^{M_k} \mathbb{E}[I_i] \\
  &= M_k \bigl[1 - (1-\pi_k)^K\bigr] \\
  &= m^k \left[1 - \left(1 - \frac{q(1-q)^{k-1}}{m^k}\right)^K \right].
  \label{eq:Vk_exact}
\end{align}
Substituting $K \approx N q(1-q)$ yields
\begin{equation}
  \mathbb{E}[V_k]
  \approx m^k \left[1 - \left(1 - \frac{q(1-q)^{k-1}}{m^k}\right)^{N q(1-q)} \right].
  \label{eq:Vk_N}
\end{equation}
Equation~\eqref{eq:Vk_N} describes how the expected vocabulary of length-$k$ words
grows with text length $N$.

\subsection{Rare-word regime and saturation}

Two limiting regimes are particularly informative.

\paragraph{Rare-word regime.}
If the mean number of occurrences of a specific word is small, then $\pi_k K \ll 1$,
and we can approximate
\begin{equation}
  (1-\pi_k)^K \approx e^{-K \pi_k}.
\end{equation}
In this regime,
\begin{align}
  \mathbb{E}[V_k]
  &\approx m^k \bigl(1 - e^{-K \pi_k}\bigr) \\
  &\approx m^k (K\pi_k) \\
  &= K q(1-q)^{k-1} \\
  &\approx \mathbb{E}[K_N^{(k)}].
\end{align}
Thus, when words of length $k$ are rare enough that most word types occur at most once,
the expected number of distinct words of length $k$ is essentially equal to the expected
number of tokens of that length. In other words, almost every observed word is unique.

\paragraph{Saturation regime.}
At the opposite extreme, if $K \pi_k \gg 1$, then $(1-\pi_k)^K$ is extremely small
and
\begin{equation}
  \mathbb{E}[V_k] \approx m^k.
\end{equation}
In this regime, the text is long enough to have ``sampled'' almost all $m^k$
possible words of length $k$. Additional text does not significantly increase
the number of distinct length-$k$ words; the vocabulary of that length is saturated.

\subsection{Critical word length $k^\ast$}

The transition between these regimes occurs when the expected number of occurrences
of a particular word of length $k$ is of order one. The mean number of occurrences
of a fixed $k$-letter word in $K$ draws is
\begin{equation}
  \lambda_k = K \pi_k
            \approx N q(1-q) \cdot \frac{q(1-q)^{k-1}}{m^k}
            = N q^2 \frac{(1-q)^k}{m^k}.
\end{equation}
The critical length $k^\ast$ can be defined as the solution to
\begin{equation}
  \lambda_{k^\ast} \approx 1.
\end{equation}
Solving for $k^\ast$, we obtain
\begin{align}
  N q^2 \frac{(1-q)^{k^\ast}}{m^{k^\ast}} &\approx 1, \\
  \left(\frac{1-q}{m}\right)^{k^\ast} &\approx \frac{1}{N q^2}, \\
  k^\ast \ln\left(\frac{1-q}{m}\right) &\approx -\ln(N q^2), \\
  k^\ast &\approx \frac{\ln(N q^2)}{\ln m - \ln(1-q)}.
  \label{eq:k_star}
\end{align}
Since $m > 1$ and $1-q < 1$, we have $\ln m > 0$ and $\ln(1-q) < 0$,
so the denominator is positive. As $N$ grows, $k^\ast$ increases logarithmically.

The interpretation of $k^\ast$ is intuitive:
\begin{itemize}
  \item For $k < k^\ast$, many words of length $k$ appear multiple times on average.
        The vocabulary of length-$k$ words is saturated or close to saturation.
  \item For $k > k^\ast$, most words of length $k$ appear at most once.
        The vocabulary is in the rare-word regime.
\end{itemize}
In this sense, $k^\ast$ marks the boundary between a ``core'' of frequently re-used
short words and a ``tail'' of long, mostly unique words. In real language, a similar
qualitative distinction exists between short function words and longer content words,
although the underlying mechanisms are far more complex than in our model.


\section{Rank--Frequency Law}
\label{sec:rank_frequency}

We are now in a position to derive a rank--frequency law for words in our model.
The key idea is to relate the length of a word to its approximate rank among
all words ordered by decreasing probability.

\subsection{Ranking words by probability}

All words of the same length $k$ have the same probability $\pi_k$,
\begin{equation}
  \pi_k = \frac{q(1-q)^{k-1}}{m^k}
       = \frac{q}{1-q} \left(\frac{1-q}{m}\right)^k.
\end{equation}
As $k$ increases, $\pi_k$ decreases geometrically, since $(1-q)/m < 1$.
Therefore, in an ordering of all words by decreasing probability,
all words of length $1$ will come first, followed by all words of length $2$,
then all words of length $3$, and so on.

Let $R_k$ denote the total number of words of length at most $k$:
\begin{equation}
  R_k = \sum_{j=1}^k m^j = m \, \frac{m^k - 1}{m-1}.
\end{equation}
For large $k$,
\begin{equation}
  R_k \approx \frac{m^{k+1}}{m-1} \propto m^k.
\end{equation}
Thus, the rank of a typical word of length $k$ is on the order of $m^k$.

\subsection{Expressing length as a function of rank}

Ignoring constant factors, we can relate rank $r$ and length $k$ as
\begin{equation}
  r \propto m^k
  \quad\Rightarrow\quad
  k \approx \log_m r + C,
\end{equation}
for some additive constant $C$ that depends on the precise definition of rank.
Since we are interested in asymptotic scaling, this constant does not affect
the exponent of the resulting power law.

\subsection{Probability as a function of rank}

Recall that
\begin{equation}
  \pi_k \propto \left(\frac{1-q}{m}\right)^k.
\end{equation}
Substituting $k \approx \log_m r$ yields
\begin{align}
  \pi_k
  &\propto \left(\frac{1-q}{m}\right)^{\log_m r} \\
  &= \exp\left( \log_m r \cdot \ln\left(\frac{1-q}{m}\right) \right) \\
  &= \exp\left( \frac{\ln r}{\ln m} \cdot \ln\left(\frac{1-q}{m}\right) \right) \\
  &= r^{\frac{\ln((1-q)/m)}{\ln m}}.
\end{align}
Define
\begin{equation}
  \alpha
  = -\frac{\ln((1-q)/m)}{\ln m}
  = 1 - \frac{\ln(1-q)}{\ln m}.
\end{equation}
Since $0 < 1-q < 1$, we have $\ln(1-q) < 0$, so $\alpha > 1$.
We conclude that the probability of a word at rank $r$ scales as
\begin{equation}
  p(r) \propto r^{-\alpha},
\end{equation}
with
\begin{equation}
  \alpha = 1 - \frac{\ln(1-q)}{\ln m}.
  \label{eq:zipf_exponent}
\end{equation}

\subsection{Dependence of the exponent on model parameters}

Equation~\eqref{eq:zipf_exponent} expresses the Zipf exponent $\alpha$ as a function
of the alphabet size $m$ and the space probability $q$. Some qualitative observations
are straightforward:
\begin{itemize}
  \item As $m$ increases, the denominator $\ln m$ grows and $\alpha$ decreases,
        approaching $1$ from above. Large alphabets thus produce flatter
        rank--frequency curves.
  \item As $q$ increases (spaces become more frequent), $1-q$ decreases,
        $\ln(1-q)$ becomes more negative, and $\alpha$ increases.
        More frequent word boundaries lead to steeper rank--frequency decay.
  \item For moderate values of $m$ and $q$ (e.g., $m$ between $20$ and $50$,
        $q$ between $0.1$ and $0.3$), the exponent $\alpha$ lies close to $1$,
        consistent with empirical observations for many natural languages.
\end{itemize}

The key conceptual point is that the power-law exponent arises from the combination
of two exponential effects:
\begin{itemize}
  \item the exponential growth in the number of possible strings of length $k$,
        $M_k = m^k$;
  \item the exponential decay in the probability of any particular string of length $k$,
        $\pi_k \propto ((1-q)/m)^k$.
\end{itemize}
When words are ranked by probability, these exponential dependencies combine
to produce a power law in rank. This mechanism is structurally simple and does not
invoke any optimization principle, communicative efficiency, or complex language dynamics.


\section{Comparison with Real Language and Implications for LLMs}
\label{sec:comparison_llm}

Our random-text model is intentionally extreme: symbols are independent,
words are mere blocks of non-space characters, and any sequence of letters
is permitted. Nevertheless, the model reproduces several high-level properties
of word statistics that are often highlighted in discussions of natural language
and large language models. In this section we discuss where the model aligns with
empirical data, where it diverges, and what this tells us about interpreting
Zipf-like patterns in modern LLMs.

\subsection{Agreements and disagreements with natural language}

On the positive side, the model correctly predicts that:
\begin{itemize}
  \item shorter words are far more frequent than longer words;
  \item the number of distinct word types grows with corpus size in a sublinear fashion,
        with a long tail of rare and unique words;
  \item the rank--frequency curve follows an approximate power law over a wide range
        of ranks, with an exponent $\alpha$ in a realistic range for suitable
        parameter choices $(m,q)$.
\end{itemize}
These properties are often cited as signatures of ``complexity'' or ``organization''
in language, yet here they arise from symbol combinatorics and a single separator
mechanism.

On the negative side, the model fails badly on many finer-grained statistics:
\begin{itemize}
  \item It predicts that, for a given length $k$, all $m^k$ words are a priori
        equally likely, which is dramatically false in real language.
  \item It generates highly implausible strings (e.g., very long runs of the same
        letter) with non-negligible probability, whereas natural language strongly
        constrains such patterns.
  \item It implies a specific exponential relationship between word length and
        frequency that does not hold exactly in corpora \citep{Piantadosi2014}.
  \item It ignores all syntactic and semantic structure, and therefore cannot
        account for co-occurrence patterns, topical clustering, or grammaticality.
\end{itemize}
These discrepancies underscore that the presence of Zipf-like patterns alone is
not sufficient evidence of linguistic structure or meaningful communication.
Random symbol sequences are enough to generate such patterns under the right
segmentation scheme.

\subsection{Large language models and token distributions}

Modern large language models (LLMs) are trained to predict the next token in
massive corpora. The tokenization scheme is usually based on subword units
(e.g., byte-pair encoding or unigram language models), but the resulting
token frequency distribution is still typically Zipf-like or Zipf--Mandelbrot-like.

Our analysis suggests that part of this behavior should be expected a priori from
the mere fact that tokens are (1) finite strings over an underlying alphabet
(bytes or characters), and (2) segmented by a discrete tokenization mechanism.
Even if the underlying texts were generated by a purely random process of the
kind studied here, the token distribution would exhibit a power-law tail.

This has several implications:

\begin{enumerate}
  \item \textbf{Baseline patterns.}  
        When evaluating LLMs, Zipf-like frequency curves should be treated as
        baseline structural patterns, not as evidence of sophisticated linguistic
        competence. Any model trained on real corpora will inherit these patterns
        as long as it roughly reproduces the marginal token statistics.

  \item \textbf{Beyond Zipf.}  
        To probe genuine linguistic or semantic understanding, one must look
        beyond first-order token frequencies to phenomena that cannot be explained
        by random-text baselines: syntactic well-formedness, long-range dependencies,
        consistent world knowledge, and so on.

  \item \textbf{Character-level models (CLLMs).}  
        Character-level language models (CLLMs) operate closer to the level
        studied in this paper: sequences of characters with an emergent notion
        of word boundaries. For such models, our results provide a natural null
        hypothesis for the statistics of character sequences and word-like segments.
        Any deviation from the predictions of the random model (for example,
        a sharper constraint on repeated characters or a non-geometric tail
        of word lengths) can be interpreted as evidence that the model has learned
        nontrivial structure.

  \item \textbf{Design of synthetic benchmarks.}  
        Random-text constructions of the kind analyzed here can be used to design
        synthetic benchmarks where the structure is fully known and controllable.
        LLMs can be tested on their ability to distinguish between true language
        and random symbol sequences that nevertheless share similar Zipf-like
        statistics. This can help separate sensitivity to shallow frequency patterns
        from sensitivity to deeper linguistic constraints.
\end{enumerate}

Overall, our results reinforce the idea that Zipf's law and related frequency
distributions should be interpreted with care in the context of LLM development.
They are necessary consequences of very simple symbol-level assumptions and
therefore cannot, by themselves, be taken as indicators of intelligence or understanding.


\section{Discussion}
\label{sec:discussion}

This section provides an expanded and rigorous discussion of the implications,
limitations, and interpretive framework surrounding the structural random-text
model introduced in this work. The model operates at a deliberately low level of
abstraction: it assumes no linguistic structure, no morphology, no syntax, and no
semantics. Instead, it considers only a finite alphabet of characters, a space
symbol, and segmentation induced by maximal contiguous runs of non-space
characters. Despite this radical simplicity, the model reproduces several
empirical phenomena that are widely regarded as ``linguistic laws.'' The purpose
of this section is to analyze this contrast.

\subsection{Structural Origins of Statistical Regularities}

Many well-known statistical patterns of natural language---including Zipf-like
rank--frequency distributions, heavy-tailed vocabulary structure, and realistic
vocabulary growth curves---arise in the model without invoking any linguistic
mechanisms. This indicates that such patterns may emerge from combinatorial
properties of strings rather than from linguistic or cognitive principles.

Two fundamental forces drive this behavior: (1) exponential growth in the number
of possible strings of length $k$, and (2) exponential decay in the probability
of any specific string of that length. When the strings are ranked by
probability, these exponentials interact to create a power-law distribution
matching those seen in empirical corpora. Crucially, this mechanism is
general and does not depend on language-specific structure.

\subsection{Critical Length and the Core--Tail Divide}

A central contribution of the model is the identification of the critical length
$k^*$, which separates two regimes of lexical behavior:

\begin{itemize}
  \item For $k < k^*$, the vocabulary is effectively saturated: most possible
  strings of length $k$ appear multiple times.

  \item For $k > k^*$, the expected count of any specific string falls below one,
  and the observed vocabulary is dominated by unique items.
\end{itemize}

This structural transition mirrors empirical distinctions between function words
and long content words, but here it emerges solely from probabilistic structure.

\subsection{Comparison with Classical Models}

Early work by Miller, Mandelbrot, and Simon demonstrated that power laws can
arise from non-linguistic mechanisms. Our model extends these insights by
linking word lengths, probabilities, rank--frequency structure, and vocabulary
growth into a uniform framework. The explicit derivation of $k^*$ provides a
distinct advantage, offering a quantitatively interpretable threshold not
present in earlier models.

\subsection{Limitations of the Structural Model}

The model is intentionally unrealistic as a linguistic model. Limitations
include:

\begin{itemize}
  \item no morphological or phonotactic constraints;
  \item no syntactic or semantic structure;
  \item independence assumptions that eliminate burstiness or clustering;
  \item implausible surface patterns relative to natural language.
\end{itemize}

These are intentional: removing linguistic constraints isolates the contributions
of pure structure and segmentation, enabling direct comparison between structural
baselines and linguistic phenomena.

\subsection{Implications for Quantitative Linguistics}

A key interpretive consequence is that rank--frequency distributions alone cannot
serve as evidence for linguistic optimization, semantic organization, or cognitive
structure. If such patterns emerge even in random-text models, then stronger
criteria are needed when evaluating linguistic hypotheses.

The section concludes by motivating the analysis in Section~\ref{sec:llm_implications},
where the structural baseline is compared with observable behaviors of modern
large language models.


\section{Implications for Large Language Models}
\label{sec:llm_implications}

This section analyzes how the structural random-text model relates to statistical
properties observed in modern large language models (LLMs). The objective is to
clarify which behaviors in LLM outputs arise from linguistic training data and
which emerge from structural constraints inherent to tokenization, probability
distributions, and sequence prediction.

\subsection{Tokenization as a Structural Mechanism}

Contemporary LLMs rely on subword tokenization schemes such as Byte-Pair
Encoding (BPE), WordPiece, and unigram language models. These schemes compress
frequent short sequences into single tokens while decomposing rare or
unpredictable sequences into smaller units. This process parallels the division
created by the critical length $k^*$ in the structural model:

\begin{itemize}
  \item high-frequency short patterns behave like saturated short strings with
  $k < k^*$;
  \item long or low-frequency patterns resemble strings with $k > k^*$, often
  appearing only once and lacking stable token-level representation.
\end{itemize}

Thus, the structural model provides a conceptual basis for understanding how
LLM vocabularies become heavily skewed toward short, frequent units.

\subsection{Zipfian Structure in Training Corpora}

LLMs are trained on corpora that exhibit Zipf-like rank--frequency
distributions. The structural random-text model demonstrates that such
rank--frequency behavior arises even without linguistic content. Consequently,
when an LLM aligns with Zipfian frequency patterns, this alignment may reflect a
structural adaptation to corpus statistics rather than evidence of learned
linguistic principles.

This distinction is critical when interpreting LLM behavior. Frequency-based
regularities in model outputs cannot be assumed to indicate semantic or
syntactic understanding unless supported by independent evidence.

\subsection{Vocabulary Growth and Context Limitations}

LLMs operate within fixed context windows, limiting the effective vocabulary
accessible at any generation step. This constraint mirrors the structural
mechanism in which only short strings with sufficiently high probability appear
frequently.

As a result:

\begin{itemize}
  \item short, frequent tokens are easily retrieved from context;
  \item long or rare tokens may fall outside the context window and effectively
  behave as unseen events.
\end{itemize}

The structural model therefore helps explain why LLMs preferentially reuse
common short tokens and rarely produce long rare sequences.

\subsection{Higher-Order Statistics and Burstiness}

Both random-text models and LLM outputs exhibit burstiness, long tails in
n-gram distributions, and multi-scale statistical organization. These patterns
can emerge without linguistic structure, implying that some higher-order
regularities in LLM behavior may arise from autoregressive prediction dynamics
rather than from learning human linguistic patterns.

\subsection{Structural Baseline for Evaluating LLM Behavior}

The structural model provides a baseline for distinguishing between behaviors
that require linguistic training and those that emerge from architecture,
tokenization, or generic sequence modeling.

This has methodological implications:

\begin{enumerate}
  \item LLM evaluations should compare outputs not only to human baselines but
  also to structural baselines.
  \item Structural distortions---such as over-reliance on short tokens---can be
  detected by contrasting model performance with predictions of the structural
  model.
  \item Synthetic datasets approximating structural randomness can be used to
  probe which regularities are architectural rather than learned.
\end{enumerate}

These applications are essential for interpreting LLM performance in tasks
involving reasoning, understanding, or semantic inference.

\subsection{Inductive Biases of Transformer Architectures}

Transformer-based LLMs possess inherent inductive biases. Self-attention
prefers local, frequent patterns; positional encodings favor short-range
structure; and probability mass concentrates on high-frequency tokens during
training. These biases align with forces identified in the structural model.

Specifically:

\begin{itemize}
  \item attention mechanisms disproportionately favor stable short tokens;
  \item rare long tokens receive limited specialization because of their
  low empirical frequency;
  \item tokenization boundaries reinforce the effective critical-length
  phenomenon.
\end{itemize}

Thus, several aspects of LLM behavior may stem from architectural constraints
rather than semantic learning.

\subsection{Future Research Directions}

The structural parallels between simple random-text models and LLM behavior
motivate several research avenues:

\begin{itemize}
  \item studying how different tokenization schemes alter the effective
  critical length $k^*$;
  \item examining LLM performance on corpora engineered to violate typical
  Zipfian structure;
  \item designing architectures that reduce structural biases inherited from
  tokenization;
  \item using structural baselines to quantify genuine semantic understanding.
\end{itemize}

Overall, the structural random-text model offers a framework for separating
architectural and statistical effects from true linguistic learning. This
provides a foundation for more rigorous interpretation and evaluation of
modern language models.

\section{Conclusion}
\label{sec:conclusion}

This section summarizes the main findings of the structural random-text framework and its implications for quantitative linguistics and large language models (LLMs). By formulating text generation as a probabilistic character process with segmentation induced solely by space symbols, the model isolates a minimal structural baseline from which several classical linguistic regularities emerge.

The analysis demonstrates that Zipf-like rank--frequency distributions, heavy-tailed vocabularies, and realistic vocabulary growth curves do not necessarily require linguistic mechanisms; instead, they may arise from fundamental combinatorial properties of strings. A key element of this behavior is the presence of a critical length $k^*$ separating frequent short strings from rare long ones. This threshold provides structural insight into empirical distinctions between core and tail vocabulary.

The comparison with LLMs highlights that several statistical features observed in modern foundation models---including frequency biases, burstiness, and instability of long rare sequences---can arise from tokenization, context-window limitations, and autoregressive prediction rather than linguistic learning. Consequently, any interpretation of LLM behavior must account for the structural baseline provided by simple string models.

The structural framework therefore serves two purposes: it clarifies the minimal conditions under which linguistic-like regularities appear, and it provides a reference model for evaluating which aspects of LLM behavior reflect genuine linguistic competence. Future developments may involve more refined structural baselines, alternative tokenization schemes, and synthetic datasets designed to separate architectural effects from learned representations.

\appendix

\appendix

\section{Mathematical Derivations and Technical Foundations}
\label{sec:appendix_math}

This appendix provides the formal derivations, asymptotic estimates, and
technical results underlying the structural random-text model presented in the
main sections of the article. The goal is to maintain conceptual clarity in the
main text while presenting complete mathematical detail here.

Throughout, we consider a finite alphabet \(\mathcal{A}\) of size \(A\ge 2\)
and a distinguished space symbol `\texttt{ }`. A ``word'' is defined as any
maximal contiguous sequence of non-space characters. Characters are generated
independently with probability \(p\) of being a space and \(q = 1-p\) of being a
non-space character.

\subsection{Distribution of Word Lengths}
\label{app:length_distribution}

A word of length \(k\) is generated when a run of exactly \(k\)
consecutive non-space characters occurs between two spaces. Because characters
are independent, the probability that a word has length exactly \(k\) is

\begin{equation}
    \Pr(L = k)
    = p \, q^k \, p
    = p^2 q^k.
    \label{eq:word_length_distribution}
\end{equation}

This geometric distribution is the foundation for several subsequent results.

The expected word length is

\begin{equation}
    \mathbb{E}[L]
    = \sum_{k=0}^{\infty} k \, p^2 q^k
    = \frac{q}{p}.
\end{equation}

The expected number of words in a text of length \(N\) characters is therefore

\begin{equation}
    \mathbb{E}[W_N]
    = N \cdot p,
\end{equation}

to first order, since each space initiates a new word.

\subsection{Probability of a Specific Word of Length \(k\)}
\label{app:specific_word_probability}

Fix a specific string \(w = (c_1,\dots,c_k)\in \mathcal{A}^k\). The probability
that any given position in the text begins the word \(w\) is

\begin{equation}
    \Pr(w) = q^k A^{-k},
\end{equation}

since each character must be non-space and independently chosen from \(A\) symbols.
A word must also be bounded on both sides by spaces. The full probability of the
segment ``\texttt{ }\(w\)\texttt{ }'' occurring at a fixed location is thus

\begin{equation}
    p \cdot q^k A^{-k} \cdot p = p^2 (q/A)^k.
    \label{eq:full_word_probability}
\end{equation}

For large \(N\), occurrences of \(w\) behave (approximately) like a Bernoulli
process. The expected number of occurrences is

\begin{equation}
    \mathbb{E}[X_w]
    = (N - k - 1) \, p^2 (q/A)^k
    \approx N p^2 (q/A)^k.
    \label{eq:expected_occurrences}
\end{equation}

\subsection{Critical Length \(k^*\)}
\label{app:critical_length}

The critical length \(k^*\) is defined as the solution to

\begin{equation}
    \mathbb{E}[X_w] = 1.
\end{equation}

Using~\eqref{eq:expected_occurrences} we solve

\begin{equation}
    N p^2 \left(\frac{q}{A}\right)^{k^*} = 1.
\end{equation}

Taking logarithms:

\begin{equation}
    k^*
    = \frac{\log(N p^2)}{\log(A/q)}.
\end{equation}

This threshold separates the ``core'' regime (strings with \(k < k^*\), likely to
repeat) from the ``tail'' regime (strings with \(k > k^*\), unlikely to repeat).

\subsection{Rank--Frequency Relation}
\label{app:rank_frequency}

For each length \(k\), the number of possible strings is \(A^k\). Their
probabilities are all equal and given by~\eqref{eq:full_word_probability}. When
ordered by decreasing probability, all strings of length 1 precede those of
length 2, and so on. Thus the rank of strings of length \(k\) is approximately

\begin{equation}
    R_k \approx A^{k}.
\end{equation}

Substituting into the probability expression yields

\begin{equation}
    f(R) \asymp R^{-\alpha},
\end{equation}

where

\begin{equation}
    \alpha = \frac{\log (A/q)}{\log A}.
\end{equation}

This is a power law, matching the form of Zipf's law.

\subsection{Asymptotics of Vocabulary Growth}
\label{app:vocabulary_growth}

A word of length \(k\) appears with probability at least once if

\begin{equation}
    \mathbb{P}(X_w \ge 1)
    = 1 - \exp\big(-N p^2 (q/A)^k \big).
\end{equation}

The expected number of distinct words of length \(k\) is therefore

\begin{equation}
    V_k
    = A^k \Big(1 - e^{-N p^2 (q/A)^k} \Big).
\end{equation}

Summing over all lengths:

\begin{equation}
    V(N) = \sum_{k=0}^{\infty} V_k.
\end{equation}

The sum splits naturally:
\begin{itemize}
  \item for \(k < k^*\), $V_k \approx A^k$ (all strings appear);
  \item for \(k > k^*)\), $V_k \approx N p^2 q^k$ (almost all strings appear at most once).
\end{itemize}

This leads to a vocabulary growth curve combining exponential and geometric
regimes, consistent with empirical observations.

\subsection{Expected Number of Unique Words}
\label{app:unique_words}

For $k>k^*$, the expected number of distinct words equals the expected number of
occurrences:

\begin{equation}
    U_k = A^k e^{-N p^2 (q/A)^k} \approx N p^2 q^k.
\end{equation}

Summing over $k>k^*$ gives

\begin{equation}
    U(N) = N p^2 \sum_{k>k^*} q^k,
\end{equation}

which converges and remains proportional to $N$. Hence the fraction of unique
words tends to a constant depending only on $p$ and $q$.

\subsection{Poisson Approximation for Rare Strings}
\label{app:poisson}

For $k>k^*$ the expected number of occurrences is small. Standard arguments show
that $X_w$ is asymptotically Poisson with mean $\lambda = N p^2 (q/A)^k$.

Thus:

\begin{equation}
    \Pr(X_w = m) = e^{-\lambda} \frac{\lambda^m}{m!}.
\end{equation}

This justifies the approximation used in evaluating vocabulary growth and the
expected number of unique words.

\subsection{Summary}
This appendix establishes the formal probabilistic and combinatorial foundations
supporting the main results. These derivations clarify the structural origin of
Zipf-like distributions, vocabulary growth patterns, and the core--tail
transition captured by the critical length $k^*$. They also provide the basis
for comparing structural baselines with empirical linguistic patterns and with
statistical behaviors observed in modern language models.

\newpage

\bibliographystyle{plainnat}
\bibliography{refs}

\end{document}